\documentclass[conference]{IEEEtran}
\IEEEoverridecommandlockouts
\usepackage{cite}
\usepackage{amsmath,amssymb,amsfonts}
\usepackage{algorithmic}
\usepackage{graphicx}
\usepackage{textcomp}
\usepackage{xcolor}

\usepackage{stfloats}
\usepackage{amsmath}    
\usepackage{amssymb}    
\usepackage{amsfonts}   
\usepackage{graphicx}   
\usepackage{booktabs}   
\usepackage{multirow}   

\usepackage[T1]{fontenc}
\usepackage[utf8]{inputenc}
\usepackage{authblk}

\newcommand{\RNum}[1]{\uppercase\expandafter{\romannumeral #1\relax}}

\def\BibTeX{{\rm B\kern-.05em{\sc i\kern-.025em b}\kern-.08em
    T\kern-.1667em\lower.7ex\hbox{E}\kern-.125emX}}
\begin{document}

\title{DDNet: A Dual-Stream Graph Learning and Disentanglement Framework for Temporal Forgery Localization}

\author[1]{Boyang Zhao}
\author[1*]{Xin Liao\thanks{Corresponding author: xinliao@hnu.edu.cn}}
\author[2]{Jiaxin Chen}
\author[1]{Xiaoshuai Wu}
\author[1]{Yufeng Wu}

\affil[1]{College of Cyber Science and Technology, Hunan University, Changsha, China}
\affil[2]{School of Computer Science and Technology, Changsha University of Science \& Technology, Changsha, China}

\maketitle

\begin{abstract}

The rapid evolution of AIGC technology enables misleading viewers by tampering mere small segments within a video, rendering video-level detection inaccurate and unpersuasive. Consequently, temporal forgery localization (TFL), which aims to precisely pinpoint tampered segments, becomes critical. However, existing methods are often constrained by \emph{local view}, failing to capture global anomalies. To address this, we propose a \underline{d}ual-stream graph learning and \underline{d}isentanglement framework for temporal forgery localization (DDNet). By coordinating a \emph{Temporal Distance Stream} for local artifacts and a \emph{Semantic Content Stream} for long-range connections, DDNet prevents global cues from being drowned out by local smoothness. Furthermore, we introduce Trace Disentanglement and Adaptation (TDA) to isolate generic forgery fingerprints, alongside Cross-Level Feature Embedding (CLFE) to construct a robust feature foundation via deep fusion of hierarchical features. Experiments on ForgeryNet and TVIL benchmarks demonstrate that our method outperforms state-of-the-art approaches by approximately 9\% in AP@0.95, with significant improvements in cross-domain robustness.
\end{abstract}


\section{Introduction}
\label{sec:intro}

\begin{figure}[t]
  \centering
  \includegraphics[width=\linewidth]{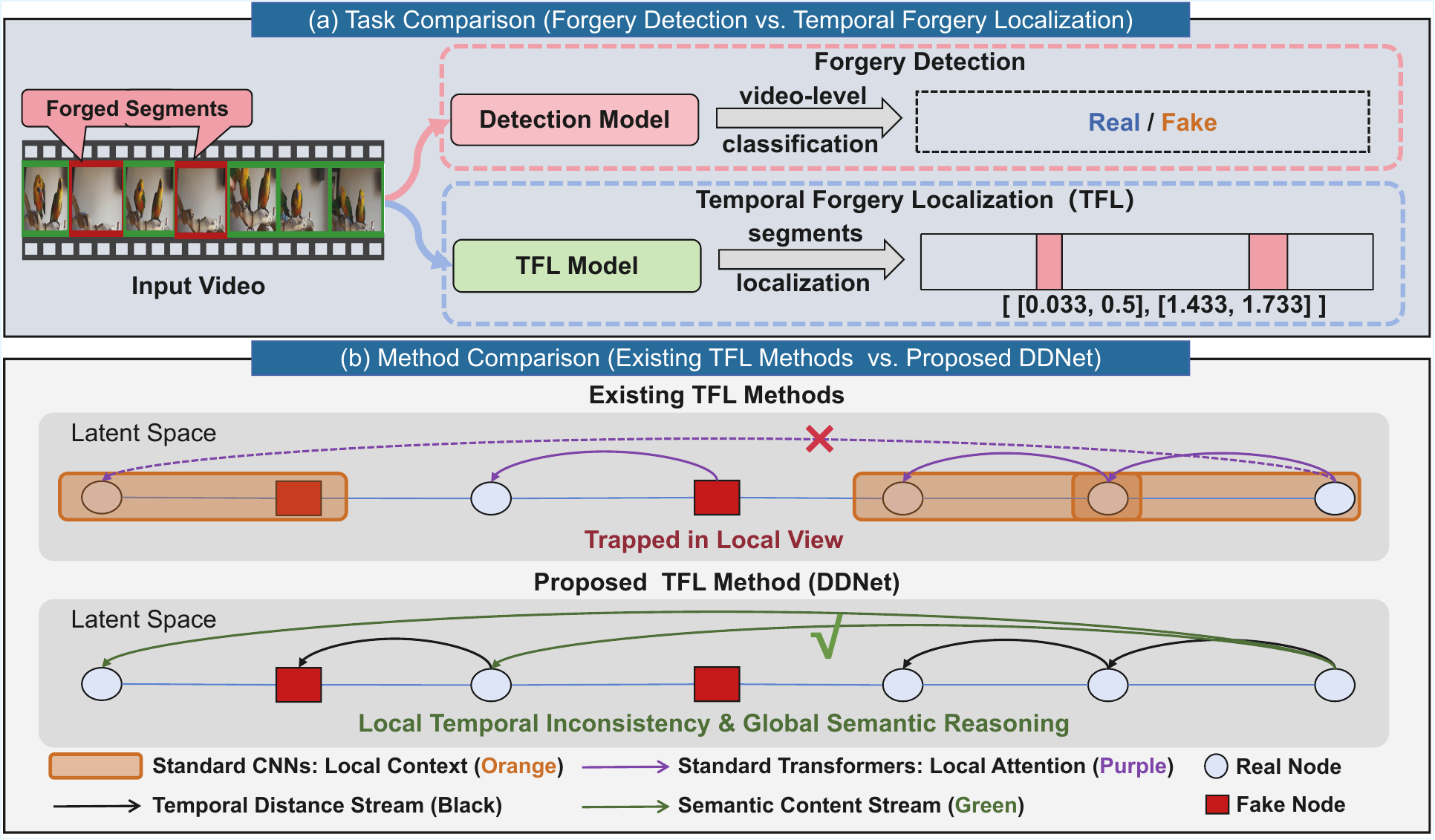} 
  \caption{{Illustration of research background and motivation.} {(a)} Comparison between forgery detection and temporal forgery localization (TFL). Forgery detection outputs a single video-level label (Real/Fake), often overlooking brief forgery segments. In contrast, TFL aims to precisely determine the location of forged segments.
{(b)} Comparison between existing TFL methods and proposed DDNet. Existing TFL methods are often constrained by \emph{local view}, failing to capture long-range dependencies between distant nodes. Our proposed DDNet overcomes this by introducing Dual-Stream Graph Learning (DSGL). By coordinating the \emph{Temporal Distance Stream} and \emph{Semantic Content Stream}, we integrate local inconsistency with global cues, thereby achieving more precise localization.}
  \label{fig:introduction}
\end{figure}

The rapid evolution of AIGC technology has rendered video forgery increasingly sophisticated and imperceptible \cite{rossler2019faceforensics++,yan2024df40}, enabling attackers to induce malicious semantic distortion or misleading effects through the manipulation of small segments \cite{cai2023glitch}. In such scenarios where forgery traces are extremely subtle, obtaining accurate binary classification results is difficult; moreover, due to a lack of interpretability, these results often fail to provide convincing evidence\cite{miao2025ddl, zhang2023ummaformer}. Consequently, temporal forgery localization (TFL) has emerged as a critical task in video forensics \cite{cai2022you, zhang2023ummaformer}, as it provides precise timestamps of manipulated segments to support interpretable evidence analysis and significantly reduce manual inspection efforts (as shown in Fig. \ref{fig:introduction}(a)).

However, existing TFL paradigms predominantly rely on standard CNNs or Transformers, yet both suffer from \emph{local view} (as shown in Fig. \ref{fig:introduction}(b)) in the context of forgery detection. CNNs are structurally constrained by rigid receptive fields \cite{cai2022you, cai2023glitch,lin2019bmn, su2021bsn++}, which limits their ability to reference distant authentic frames for verification. Conversely, standard Transformers \cite{zhang2023ummaformer, perez2024vigo,cheng2025clformer,zhang2022actionformer} tend to bias attention weights toward highly similar local neighbors, thereby hindering the capture of critical global inconsistencies. Consequently, subtle forgery traces are often obscured by the dominant authentic background, resulting in imprecise localization.

\begin{figure*}[t!] 
    \centering
    \includegraphics[width=0.98\linewidth]{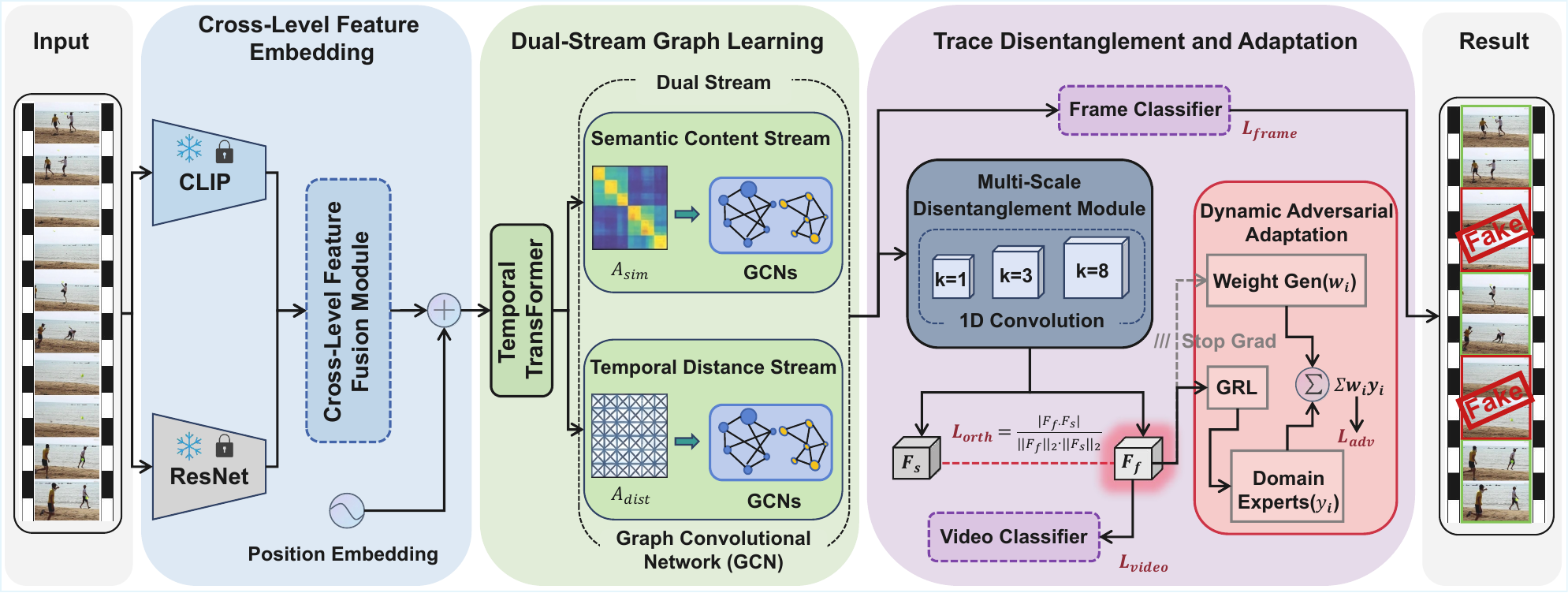} 
    
\caption{{Overview of the proposed DDNet framework.} 
The framework comprises three main components: (a) \textbf{Cross-Level Feature Embedding (CLFE):} This module extracts visual features using frozen CLIP and ResNet encoders, and subsequently fuses semantic priors with textural details via a bidirectional cross-attention mechanism to bridge the feature gap. (b) \textbf{Dual-Stream Graph Learning (DSGL):} The core module designed to overcome the \emph{local view} limitation. It integrates a \textit{Temporal Distance Stream} to capture local temporal inconsistency and a \textit{Semantic Content Stream} to perform global semantic reasoning across disjoint manipulated segments. (c) \textbf{Trace Disentanglement and Adaptation (TDA):} An auxiliary training module that imposes adversarial and orthogonal constraints to isolate generic forgery fingerprints from domain-specific noise.
Finally, the learned representations are projected to generate frame-level manipulation probabilities.}
\label{fig:framework}
\end{figure*}

To leverage global context for locating forged segments, we propose DDNet, a \underline{d}ual-stream graph learning and \underline{d}isentanglement framework. As illustrated in Fig. \ref{fig:framework}, our framework comprises three main components. Firstly, given the subtlety of forgery traces, we introduce Cross-Level Feature Embedding (CLFE) to provide a robust feature foundation. By fusing high-level semantics from CLIP \cite{radford2021learning} with low-level textures from ResNet \cite{he2016deep}, this module captures multi-grained cues essential for detecting subtle manipulations. Building on this, to resolve the core limitation of \emph{local view}, we design the Dual-Stream Graph Learning (DSGL) module. This component integrates a \emph{Temporal Distance Stream} to capture {local inconsistency} and a \emph{Semantic Content Stream} to establish {global semantic connections} between distant frames. By explicitly linking observations with distant authentic anchors, DSGL mitigates the over-smoothing phenomenon inherent in local attention mechanisms, ensuring that subtle forgery traces are not suppressed by their neighbors. Finally, to ensure the reasoning focuses on intrinsic forgery traces, we propose Trace Disentanglement and Adaptation (TDA). Acting as an implicit constraint, this auxiliary module encourages the model to discard irrelevant noise and focus on generic forgery fingerprints, thereby enhancing robustness in cross-domain scenarios.

In summary, our main contributions are as follows:
\begin{itemize}
    \item We propose DDNet, a framework featuring a Dual-Stream Graph Learning (DSGL) module. By explicitly synergizing local temporal inconsistency and global semantic reasoning, it effectively breaks the limitation of \emph{local view} and significantly improves localization precision.
    \item We introduce a Cross-Level Feature Embedding (CLFE) module and an auxiliary Trace Disentanglement and Adaptation (TDA) module. By capturing cross-level complementary features and imposing implicit constraints to focus on generic forgery fingerprints, these components jointly enhance model robustness.
    \item Extensive experiments on the ForgeryNet and TVIL benchmarks demonstrate that DDNet achieves SOTA performance, outperforming existing methods by approximately 9\% in AP@0.95, while also exhibiting significant gains in cross-domain robustness.
\end{itemize}

\section{Related Work}

\subsection{Video Forgery Detection}
Early research in video forgery detection predominantly formulated the problem as a binary classification task. Initial approaches focused on identifying image-level artifacts within individual frames \cite{rossler2019faceforensics++, li2020face}, while subsequent methods exploited temporal inconsistencies~\cite{gu2022delving,chen2022deepfake} across video clips. Despite achieving impressive accuracy on benchmarks like FaceForensics++ \cite{rossler2019faceforensics++} and DFDC \cite{dolhansky2020deepfake}, these methods are limited to providing coarse-grained labels (Real vs. Fake). As noted in recent studies \cite{miao2025ddl, zhang2023ummaformer}, binary classification lacks the interpretability required for video forensics, as it fails to pinpoint specific manipulated intervals or provide convincing evidence. This limitation has necessitated the shift toward a fine-grained task: temporal forgery localization (TFL).

\subsection{Temporal Forgery Localization}
TFL aims to precisely locate manipulated segments in a video. This objective closely mirrors that of temporal action localization (TAL), which focuses on locating the action instances within untrimmed videos. Driven by this similarity, TFL methods have largely evolved by adapting TAL architectures, primarily falling into two paradigms: boundary-matching\cite{cai2022you, cai2023glitch} and one-stage regression\cite{zhang2023ummaformer, perez2024vigo, cheng2025clformer}.

Early CNN-based adaptations~\cite{cai2022you,cai2023glitch} are constrained by rigid receptive fields, which hinder the capture of long-range dependencies. While recent Transformer-based methods\cite{zhang2023ummaformer,perez2024vigo,cheng2025clformer} offer improvements, they still suffer from a \emph{local view} limitation. Their attention mechanisms tend to overemphasize adjacent neighbors, causing critical global inconsistency cues to be drowned out by local context.

Furthermore, advanced techniques like GCNs \cite{yan2018spatial} and feature disentanglement remain underexplored in TFL. Existing methods mostly rely on entangled features, leading to semantic overfitting. To bridge this gap, DDNet integrates Dual-Stream Graph Learning (DSGL) for global reasoning and Trace Disentanglement and Adaptation (TDA) to isolate generic forgery fingerprints, effectively overcoming the limitations of \emph{local view} and weak generalization.

\begin{figure*}[t!] 
    \centering
    \includegraphics[width=0.98\linewidth]{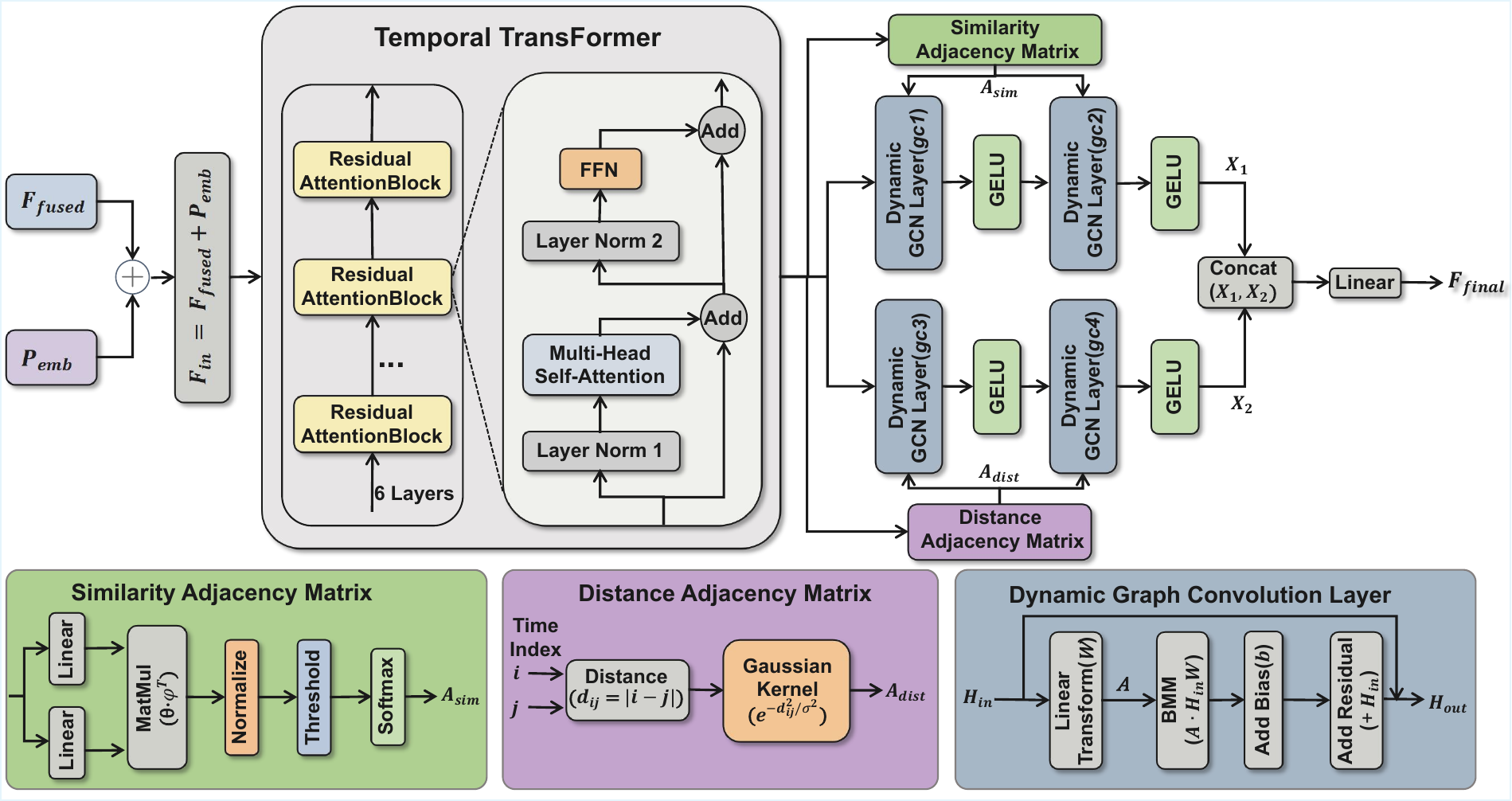} 
    \caption{{Illustration of the proposed Dual-Stream Graph Learning (DSGL) module.} 
    }
    \label{fig:DSGL}
\end{figure*}

\section{Methodology}
\label{sec:method}

\subsection{Overview}
As illustrated in Fig.~\ref{fig:framework}, {a dual-stream graph learning and
disentanglement framework for temporal forgery
localization (DDNet)} establishes a unified end-to-end framework composed of three main components:
\begin{enumerate}
    \item \textbf{Cross-Level Feature Embedding (CLFE)}: A module that constructs a robust feature foundation by leveraging visual encoders for hierarchical extraction and deeply fusing the features via a stacked bidirectional cross-attention mechanism.
    \item \textbf{Dual-Stream Graph Learning (DSGL)}: The core reasoning module that utilizes a \textit{dual-stream graph architecture}. It coordinates a \emph{Temporal Distance Stream} to enforce local inconsistency and a \emph{Semantic Content Stream} to capture global dependencies, thereby refining frame-level representations.
    \item \textbf{Trace Disentanglement and Adaptation (TDA)}: An auxiliary regularization module that employs a dynamic Mixture-of-Experts (MoE) adversarial mechanism. It enforces the backbone to learn domain-invariant forgery fingerprints by disentangling task-irrelevant noise.
\end{enumerate}
Crucially, the framework is optimized jointly, and the frame-level manipulation probabilities $\mathbf{P} = \{p_t\}_{t=1}^{T}$ are predicted directly from the DSGL output.

\subsection{Cross-Level Feature Embedding (CLFE)}
To capture both subtle textural artifacts and high-level semantic anomalies, we employ a frozen CLIP~\cite{radford2021learning} encoder and a ResNet~\cite{he2016deep} encoder to extract semantic features $\mathbf{F}_{CLIP}$ and textural features $\mathbf{F}_{Res}$, respectively. These heterogeneous features are projected to a unified dimension $D$ and fused via a stacked bidirectional cross-attention module.

In each block, let $\mathbf{H}_{CLIP}$ and $\mathbf{H}_{Res}$ denote the two intermediate feature representations. These representations interact by exchanging information, where the one serves as the \textit{Query} to retrieve context from the other (serving as \textit{Key} and \textit{Value}). Formally, the cross-modal interaction is defined as:
\begin{equation}
\label{eq:cross_attn}
    \begin{aligned}
        \mathbf{H}'_{CLIP} &= \text{MHA}(\text{LN}(\mathbf{H}_{CLIP}), \text{LN}(\mathbf{H}_{Res}), \text{LN}(\mathbf{H}_{Res})),\\
        \mathbf{H}'_{Res} &= \text{MHA}(\text{LN}(\mathbf{H}_{Res}), \text{LN}(\mathbf{H}_{CLIP}), \text{LN}(\mathbf{H}_{CLIP}))
    \end{aligned}
\end{equation}
where $\text{MHA}(\cdot)$ denotes Multi-Head Attention, $\text{LN}(\cdot)$ is Layer Normalization, and $\mathbf{H}'$ represents the updated features. This bidirectional mechanism ensures deep semantic-textural integration. The refined features are concatenated and projected to generate the fused representation $\mathbf{F}_{fused} \in \mathbb{R}^{T \times D}$.

Finally, to preserve temporal order, we introduce learnable position embeddings $\mathbf{P}_{emb} \in \mathbb{R}^{T \times D}$, which are parameterized by frame indices $t \in \{0, \dots, T-1\}$. These embeddings are added element-wise to $\mathbf{F}_{fused}$ before the sequence enters the Temporal Transformer.

\begin{table*}[t!]
\centering
\caption{Performance comparison on ForgeryNet and TVIL benchmarks. \textbf{Bold} indicates the best performance.}
\label{tab:in_domain}
\vspace{-2mm}
\resizebox{0.95\linewidth}{!}{
\begin{tabular}{l|ccc|ccc|ccc}
\toprule
\multirow{2}{*}{Method} & \multicolumn{3}{c|}{ForgeryNet (Standard)} & \multicolumn{3}{c|}{ForgeryNet  (Balanced)} & \multicolumn{3}{c}{TVIL (Inpainting)} \\
 & @0.5 & @0.75 & @0.95 & @0.5 & @0.75 & @0.95 & @0.5 & @0.75 & @0.95 \\ 
\midrule
Tridet~\cite{shi2023tridet} & - & - & -  & 21.01 & 2.63 & 0.03 & 35.58 & 6.24 & 0.43  \\
DCAN~\cite{chen2022dcan} & 71.59 & 62.02 & 32.08  & 62.80 & 50.67 & 22.91  & 79.64 & 51.59 & 0.25 \\
ActionFormer~\cite{zhang2022actionformer} & 84.49 & 78.74 & 66.10 & 72.30 & 64.90 & 54.00 & 86.30 & 83.00 & 28.20 \\
UMMAFormer~\cite{zhang2023ummaformer} & 82.30 & 78.90 & 67.70  & 74.71 & 67.43 & 54.32  & \textbf{88.70} & \textbf{84.70} & 62.40  \\ 
\textbf{Ours} & \textbf{87.59} & \textbf{83.46} & \textbf{76.70}  & \textbf{74.89} & \textbf{68.81} & \textbf{57.42}& 81.64 & 78.25 & \textbf{70.81}  \\
\bottomrule
\end{tabular}
}
\end{table*}

\subsection{Dual-Stream Graph Learning (DSGL)}
To address the limitations of \emph{local view}, we introduce the DSGL module (Fig.~\ref{fig:DSGL}). First, the fused features combined with position embeddings $\mathbf{P}_{emb}$ are processed by a \textbf{Temporal Transformer} (6 layers) to encode sequential dependencies. The output features serve as the shared input for two parallel graph streams to capture complementary cues.

\textbf{1) Temporal Distance Stream.}
To mitigate prediction flickering, this stream utilizes a static graph ${{A}_{dist}}$ to enforce local inconsistency (layers \texttt{gc3}, \texttt{gc4}). As illustrated in Fig.~\ref{fig:DSGL}, the distance adjacency matrix $\mathbf{A}_{dist} \in \mathbb{R}^{T \times T}$ is computed via a Gaussian kernel based on the temporal interval $|i-j|$ between frames $i$ and $j$:
\begin{equation}
    \mathbf{A}_{dist}(i, j) = \exp\left(-\frac{|i - j|^2}{2\sigma^2}\right),
\end{equation}
where $\sigma$ controls the bandwidth of the local receptive field, functioning as a smoothing filter.

\textbf{2) Semantic Content Stream.}
To capture long-range semantic dependencies (layers \texttt{gc1}, \texttt{gc2}), we construct a dynamic graph based on feature similarity. Following Fig.~\ref{fig:DSGL}, input features are projected via linear layers to compute pairwise similarity, followed by normalization and hard-thresholding:
\begin{equation}
    \mathbf{A}_{sim} = \text{Softmax}\left(\text{Threshold}_\tau\left(\text{Norm}\left(\mathbf{X}\mathbf{W}_\theta (\mathbf{X}\mathbf{W}_\phi)^T\right)\right)\right),
\end{equation}

where $\mathbf{W}_\theta, \mathbf{W}_\phi$ are projection matrices, and $\tau$ is the threshold to eliminate noise. We empirically set $\tau=0.7$ for robust performance on large-scale benchmarks, as supported by the sensitivity analysis in the \textbf{Supplementary Material}.

\textbf{3) Dynamic GCN \& Fusion (in Fig.~\ref{fig:DSGL}).}
Both streams employ the Dynamic Graph Convolution Layer depicted in Fig.~\ref{fig:DSGL}. For an input $\mathbf{H}_{in}$ and adjacency $\mathbf{A}$, the output $\mathbf{H}_{out}$ is formulated as:
\begin{equation}
    \mathbf{H}_{out} = \mathbf{A}(\mathbf{H}_{in}\mathbf{W}) + \mathbf{b} + \mathbf{H}_{in},
\end{equation}
where $\mathbf{W}$ and $\mathbf{b}$ denote the weight and bias. Finally, the features from both streams are concatenated and fused via a linear projection to yield $\mathbf{F}_{final}$.

\subsection{Trace Disentanglement and Adaptation (TDA)}
To isolate generic forgery fingerprints from domain-specific noise, we introduce the TDA module (Fig.~\ref{fig:framework}), comprising two key components.

\textbf{1) Multi-Scale Disentanglement.}
We first aggregate frame features into a video-level prototype via mean pooling. This prototype is processed by multi-scale 1D convolutions ($k \in \{1, 3, 8\}$) to capture multi-scale forgery patterns. The features are then projected into two orthogonal vectors: the generic forgery feature $\mathbf{F}_{f}$ and the irrelevant feature $\mathbf{F}_{s}$. To ensure separation, we minimize their cosine similarity:
\begin{equation}
    \mathcal{L}_{orth} = \frac{1}{B} \sum_{i=1}^{B} \left| \frac{\mathbf{F}_{f}^{(i)} \cdot \mathbf{F}_{s}^{(i)}}{\|\mathbf{F}_{f}^{(i)}\| \|\mathbf{F}_{s}^{(i)}\|} \right|,
\end{equation}
where $B$ is the batch size and $\|\cdot\|$ denotes the $L_2$ norm.

\textbf{2) Dynamic Adversarial Adaptation.}
To enable robust generalization, $\mathbf{F}_{f}$ is fed for domain adaptation. Instead of a single discriminator, we employ a dynamic ensemble. First, a \textbf{Weight Generator}  (in Fig. ~\ref{fig:framework}) predicts domain relevance scores $\boldsymbol{\omega} \in \mathbb{R}^K$ (where $K$ is the number of domains). Crucially, a stop-gradient operation $\text{sg}(\cdot)$ is applied to stabilize the generator:
\begin{equation}
    \boldsymbol{\omega} = \text{Softmax}(\text{Linear}(\text{sg}(\mathbf{F}_{f}))).
\end{equation}
Simultaneously, $\mathbf{F}_{f}$ passes through a Gradient Reversal Layer (GRL)~\cite{ganin2015unsupervised} to a bank of \textbf{Domain Experts} $\{\mathcal{D}_k\}_{k=1}^K$ . The final adversarial decision is the weighted summation of expert outputs:
\begin{equation}
    O_{adv} = \sum_{k=1}^{K} \omega_k \cdot \mathcal{D}_k(\text{GRL}(\mathbf{F}_{f})).
\end{equation}

\subsection{Optimization Objective}
The framework is optimized in an end-to-end manner. The TDA module serves as an auxiliary regularization component to guide the backbone in learning robust representations. The total objective function is formulated as:
\begin{equation}
    \mathcal{L}_{total} = \mathcal{L}_{frame} + \lambda_{vid}\mathcal{L}_{video} + \lambda_{adv}\mathcal{L}_{adv} + \lambda_{orth}\mathcal{L}_{orth},
\end{equation}
where $\mathcal{L}_{frame}$ and $\mathcal{L}_{video}$ denote the binary cross-entropy losses for frame-level localization and video-level detection, respectively. {Specifically, $\mathcal{L}_{adv}$ is the cross-entropy loss computed between the aggregated adversarial output $O_{adv}$ and the ground-truth forgery domain labels.}

\begin{table*}[t!]
\centering
\caption{Performance comparison on ForgeryNet and TVIL benchmarks. \textbf{Bold} indicates the best performance.}
\label{tab:cross_domain}
\vspace{-2mm}
\resizebox{0.95\linewidth}{!}{
\begin{tabular}{l|ccc|ccc|ccc}
\toprule
\multirow{2}{*}{Method} & \multicolumn{3}{c}{ForgeryNet (Standard) $\to$ TVIL} & \multicolumn{3}{c}{ForgeryNet (Balanced) $\to$ TVIL} & \multicolumn{3}{c}{TVIL $\to$ ForgeryNet(Balanced)} \\
\cmidrule(lr){2-4} \cmidrule(lr){5-7} \cmidrule(lr){8-10}
 & @0.5 & @0.75 & @0.95 & @0.5 & @0.75 & @0.95 & @0.5 & @0.75 & @0.95 \\ 
\midrule
Tridet~\cite{shi2023tridet} &- & - & -& 1.05 & 0.03 & 0 & 0.56 & 0 & 0\\
DCAN~\cite{chen2022dcan} &1.49 &0.43 & 0 & 0.83 & 0.21 & 0 & 1.12 & 0.27 & 0\\
ActionFormer~\cite{zhang2022actionformer} &2.76 & 1.40 & 0.02 & 1.12 & 0.33 & 0 & 0.82 & 0.10 & 0\\
UMMAFormer~\cite{zhang2023ummaformer} &3.02&  1.80&  0 &  2.15&  0.98&  0 & 1.26 & 0.18 & 0 \\
    Ours (no\_TDA)&7.25&5.27&0.25 & 4.25 &3.14 & 0.02 & \textbf{1.51} & \textbf{0.56} & 0 \\
\textbf{Ours} &\textbf{11.72} &\textbf{8.25}&  \textbf{0.62}&  \textbf{7.64}&  \textbf{6.43}&\textbf{0.09}  &  1.33&  0.52 & \textbf{0}  \\
\bottomrule
\end{tabular}
}
\end{table*}

\section{Experiments}
\label{sec:experiments}

\subsection{Experimental Setup}
\textbf{Datasets and Metrics.} We evaluate DDNet on two benchmarks: (1) \textbf{ForgeryNet}~\cite{he2021forgerynet}, a large-scale TFL dataset with diverse manipulation techniques. We construct \textit{Standard} and \textit{Balanced} subsets to evaluate general and class-balanced performance, respectively (details in \textbf{Supplementary Material}). (2) \textbf{TVIL}~\cite{zhang2023ummaformer}, a specialized benchmark for {video inpainting scenes}, constructed by removing objects during random time intervals using advanced inpainting algorithms.

\textbf{Evaluation metrics.} We report Average Precision (AP) at tIoU thresholds $\{0.5, 0.75, 0.95\}$.We place particular
emphasis on AP@0.95, as it reflects the model’s capability to
perform high-level precise localization.

\textbf{Baselines.} We compare against generic TAL models (ActionFormer~\cite{zhang2022actionformer}, DCAN~\cite{chen2022dcan}) and specialized TFL frameworks (UMMAFormer~\cite{zhang2023ummaformer}). Note that VIGO~\cite{perez2024vigo} shares the visual architecture with UMMAFormer; thus, comparisons with UMMAFormer reflect the current visual-based SOTA.

\textbf{Implementation Details.} Our models are implemented in PyTorch and trained using the AdamW optimizer with an initial learning rate of $2 \times 10^{-5}$. The batch size is set to 64, with the temporal dimension fixed at $T=512$. We utilize frozen CLIP and ResNet-50 as feature extractors. Regarding the loss function, the balancing coefficients are set to $\lambda_{vid}=0.3$, $\lambda_{adv}=0.005$, and $\lambda_{orth}=1.0$.

\begin{table}[t!]
\centering
\caption{Ablation study of core components.}

\label{tab:ablation}
\resizebox{0.95\linewidth}{!}{
\begin{tabular}{cc|ccc|ccc}
\toprule
\multicolumn{2}{c|}{Input} & \multicolumn{3}{c|}{Components} & \multicolumn{3}{c}{Metrics} \\
CLIP & ResNet & CLFE & GCN & TDA & @0.5 & @0.75 & @0.95 \\ 
\midrule
\checkmark & & - & \checkmark & \checkmark & 47.00 & 35.04 & 22.84 \\
 & \checkmark & - & \checkmark & \checkmark & 58.52 & 48.90 & 26.00 \\ \midrule
\checkmark & \checkmark & & \checkmark & \checkmark & 61.85 & 49.64 & 26.75 \\
\checkmark & \checkmark & \checkmark & & \checkmark & 65.39 & 59.20 & 51.10 \\
\checkmark & \checkmark & \checkmark & \checkmark & & 71.57 & 63.82 & 53.18 \\
\checkmark & \checkmark & \checkmark & \checkmark & \checkmark & \textbf{74.89} & \textbf{68.81} & \textbf{57.42} \\ 
\bottomrule
\end{tabular}
}
\end{table}

\subsection{In-Domain Performance Analysis}
Table~\ref{tab:in_domain} summarizes the quantitative results.
On the \textbf{ForgeryNet} \emph{Standard Subset}, DDNet achieves an {AP@0.95 of 76.70\%}, surpassing the SOTA UMMAFormer by \textbf{9.0\%}.
Similarly, on the \textbf{TVIL}, which features challenging inpainting attacks, DDNet maintains superior precision (70.81\% vs. 62.40\% for UMMAFormer).
This significant margin at strict tIoU threshold validates the structural advantage of our {DSGL} module. Unlike standard Transformers constrained by local neighbors, DSGL integrates temporal distance inconsistency with semantic content reasoning, effectively enhancing localization precision.

\subsection{Cross-Domain Generalization}
Table~\ref{tab:cross_domain} reveals an \textit{asymmetric generalization} pattern.

\textbf{Transfer from ForgeryNet to TVIL.} Despite the semantic gap, DDNet achieves an AP@0.5 of {11.72\%}, a \textbf{4-fold improvement} over UMMAFormer. This success indicates that the diverse manipulation traces in the ForgeryNet effectively {encompass} the simpler inpainting traces in TVIL. Thanks to the TDA module, our encoder learns to disentangle intrinsic forgery fingerprints from irrelevant noises, allowing it to easily generalize to TVIL.

\textbf{Transfer from TVIL to ForgeryNet.} Conversely, the reverse transfer proves challenging. We attribute this to \emph{domain overfitting}: models trained on TVIL focus exclusively on specific inpainting artifacts. The resulting representations are incomplete for face forensics, as they lack the high-level semantic and biological cues (e.g., facial inconsistencies) required for sophisticated deepfakes detection.

\begin{figure}[t]
  \centering
  \includegraphics[width=\linewidth]{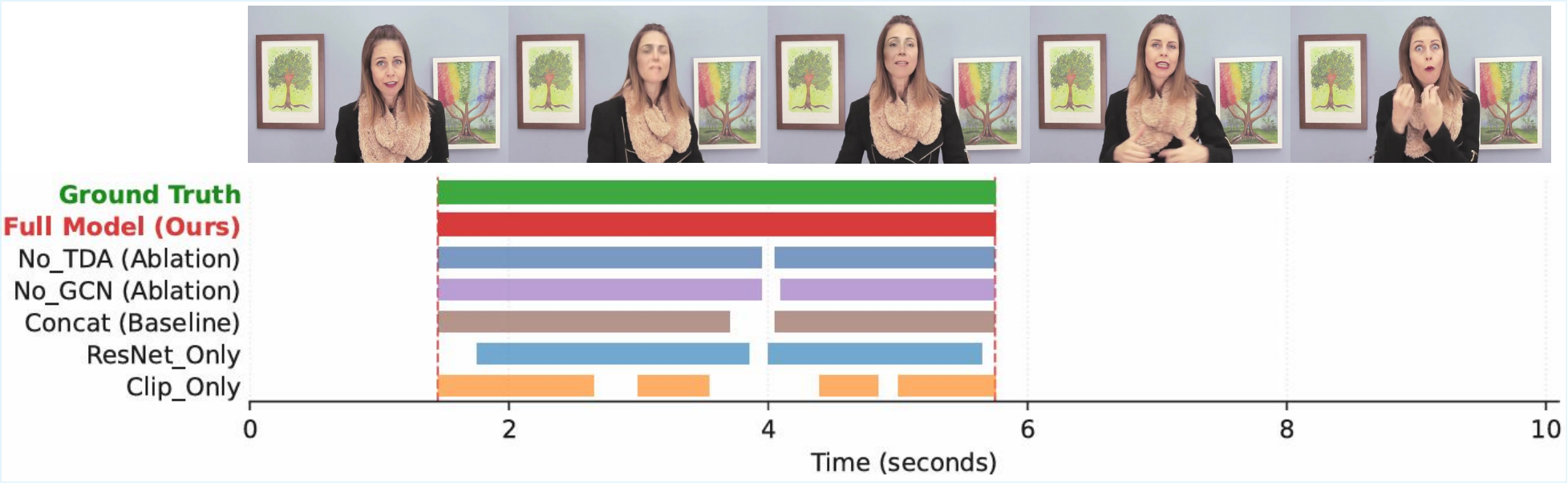}
  \caption{\textbf{Qualitative visualization.} The Ground Truth (green) shows a continuous forgery. The {ablated variants} suffer from severe \textit{prediction fragmentation}. Our full model (red) can maintain consistency, avoiding the gaps seen in ablated variants.}
  \label{fig:visualization}
\end{figure}

\subsection{Ablation Study}
We analyze component contributions on the \textbf{ForgeryNet} \emph{Balanced} subset in Table~\ref{tab:ablation}.

\textbf{1) Impact of Graph Reasoning (DSGL).} 
To isolate the contribution of GCNs, we remove the parallel GCN streams (Row 4) while retaining the Temporal Transformer. Note that removing the entire DSGL would eliminate temporal modeling capabilities. The resulting approximately 6\% drop in AP@0.95 confirms that relying solely on temporal attention is insufficient for precise localization. The GCN streams are essential for explicitly enforcing local inconsistency and global cues.

\textbf{2) Effectiveness of TDA.} 
Ablating the auxiliary TDA module (Row 5) causes a moderate drop in in-domain performance, accompanied by a degradation in cross-domain generalization (in Table ~\ref{tab:cross_domain}). This validates that TDA prevents the model from overfitting to specific video content, forcing it to learn generic forgery traces that generalize better to new domains.

\textbf{3) Necessity of CLFE.} 
Rows 1-3 demonstrate that neither single-modality backbones (lacking complementary cues) nor naive concatenation (limited by feature heterogeneity) yield optimal results. In contrast, incorporating CLFE triggers a substantial performance leap. This proves that our bidirectional cross-attention mechanism is indispensable for bridging the gap between semantic priors and textural artifacts.

\subsection{Qualitative Analysis}
\label{sec:qualitative}
Fig.~\ref{fig:visualization} visualizes the detection results on a continuous forgery sample.

\textbf{Fragmentation in Variants.} As shown in Fig.~\ref{fig:visualization}, the Ground Truth (green) represents a single continuous manipulated segment. However, the {ablated variants} suffer from severe \textit{prediction fragmentation}. Specifically, removing graph reasoning (\textit{No\_GCN}, purple) or relying on a single modality (\textit{Clip\_Only}, orange) results in disjoint fragments. This failure to maintain temporal consistency highlights the necessity of our full model.

\textbf{Precision of DDNet.} In contrast, our full model (Red) generates a highly coherent proposal that aligns precisely with the Ground Truth. DDNet effectively bridges the transitional gaps observed in the incomplete variants, eliminating fragmentation and ensuring superior localization precision.

\section{Conclusion}
\label{sec:conclusion}
In this paper, we proposed DDNet, which {significantly enhances localization precision and generalization capability} in TFL. By integrating local inconsistency with global reasoning, our DSGL module effectively overcomes the \emph{local view} limitation. Complemented by TDA and CLFE, the framework learns robust manipulation fingerprints that transfer effectively across domains. Extensive experiments on ForgeryNet and TVIL demonstrate SOTA performance, particularly in high-precision regimes. Future work will aim to bridge the remaining cross-domain semantic gaps by exploring multi-modal integration and Large Vision Models (LVMs).

\bibliographystyle{IEEEbib}
\bibliography{icme2026references}

\title{DDNet: A Dual-Stream Graph Learning and Disentanglement Framework for Temporal Forgery Localization (Supplementary Material)}

\author{Anonymous ICME submission}

\maketitle

In this supplementary material, we provide additional experimental details and analyses that were omitted from the main text due to space limitations. Specifically, we first detail the specific composition and distribution of the \emph{Standard} and \emph{Balanced} subsets used in the ForgeryNet dataset (Section~\ref{sec:dataset_details}). Furthermore, we present a comprehensive sensitivity analysis of the hyperparameter $\tau$ in the Dual-Stream Graph Learning (DSGL) module to validate our empirical parameter selection (Section~\ref{sec:sensitivity_tau}).
\section{Dataset Details}
\label{sec:dataset_details}

In our experiments, we evaluate performance on the ForgeryNet~\cite{he2021forgerynet} dataset using two distinct evaluation subsets: the \emph{Standard} subset and the \emph{Balanced} subset. These protocols are designed to assess the model's performance under different data distributions.

\begin{figure}[h]
    \centering
    \includegraphics[width=0.8\linewidth]{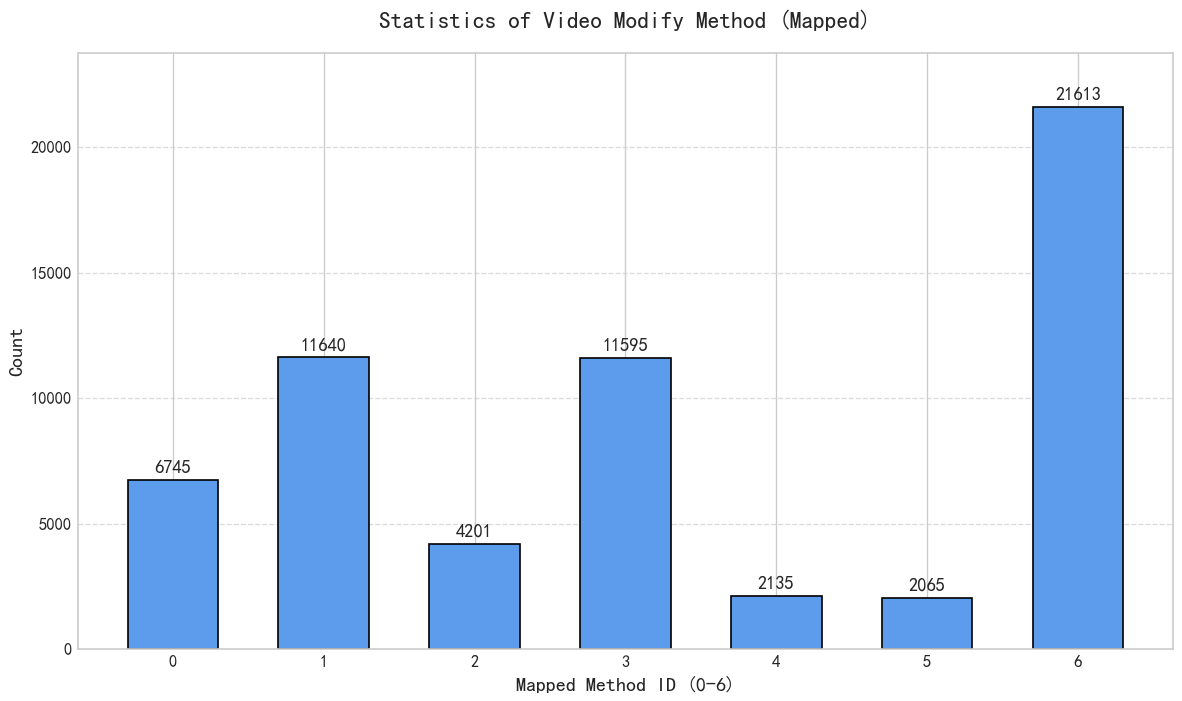} 
    \caption{{Distribution of Manipulation Methods in the \textbf{Training Set} of the \emph{Standard Subset}.} The distribution is highly imbalanced, with Method 6 accounting for a significant portion of the data, while other methods are underrepresented.}
    \label{fig:distribution}
\end{figure}

ForgeryNet~\cite{he2021forgerynet} is a comprehensive dataset characterized by a wide array of manipulation techniques and a massive scale. To comprehensively evaluate our method, we constructed two distinct subsets based on this dataset:

\begin{itemize}
    \item \textbf{Standard Subset (Natural Distribution):} 
    First, we constructed a large-scale subset by randomly sampling from the original dataset. This subset consists of \textbf{59,994} videos for training, \textbf{14,019} for testing, and \textbf{14,020} for validation. As illustrated in Fig.~\ref{fig:distribution}, the distribution of manipulation methods in the training set is highly skewed (long-tailed). \textbf{Specifically, Method ID 0 denotes real videos, while the remaining IDs correspond to videos generated by different manipulation techniques.} This reflects the natural imbalance often found in real-world scenarios.
    
    \item \textbf{Balanced Subset (Uniform Distribution):} 
    Considering the computational efficiency of experiments and the potential bias introduced by the long-tailed distribution in the Standard Subset, we further constructed a Balanced Subset. This subset comprises \textbf{5,600} training videos, \textbf{700} testing videos, and \textbf{700} validation videos. In contrast to the Standard Subset, we rigorously ensured that \textbf{each forgery method is represented by an equal number of videos} across all splits. This setting facilitates a fair assessment of the model's generalization capability across diverse forgery techniques without being dominated by majority classes.
\end{itemize}

\section{Sensitivity Analysis of Threshold {$\tau$}}
\label{sec:sensitivity_tau}

\subsection{Hyperparameter Definition and Evaluation Metric}
In the \emph{Semantic Content Stream} of our proposed DSGL module, we construct a dynamic graph to capture long-range semantic dependencies. The hyperparameter $\tau$ serves as a hard threshold applied to the normalized feature similarity matrix. 
\begin{itemize}
    \item \textbf{Meaning of $\tau$:} This parameter explicitly controls the sparsity of the semantic graph. A lower $\tau$ allows more edges to be retained, potentially introducing irrelevant or noisy connections. Conversely, a higher $\tau$ enforces stricter filtering, preserving only the most confident semantic correlations but potentially severing useful long-range dependencies.
    \item \textbf{Evaluation Metric (mAP):} To comprehensively assess the localization accuracy, we report the mean Average Precision (mAP). Specifically, this metric is calculated as the average of AP scores at three distinct Intersection over Union (IoU) thresholds:
    \begin{equation}
        \text{mAP} = \frac{\text{AP}@0.5 + \text{AP}@0.75 + \text{AP}@0.95}{3}
    \end{equation}
\end{itemize}

\subsection{Results and Discussion}
\begin{figure}[h]
    \centering
    \includegraphics[width=0.9\linewidth]{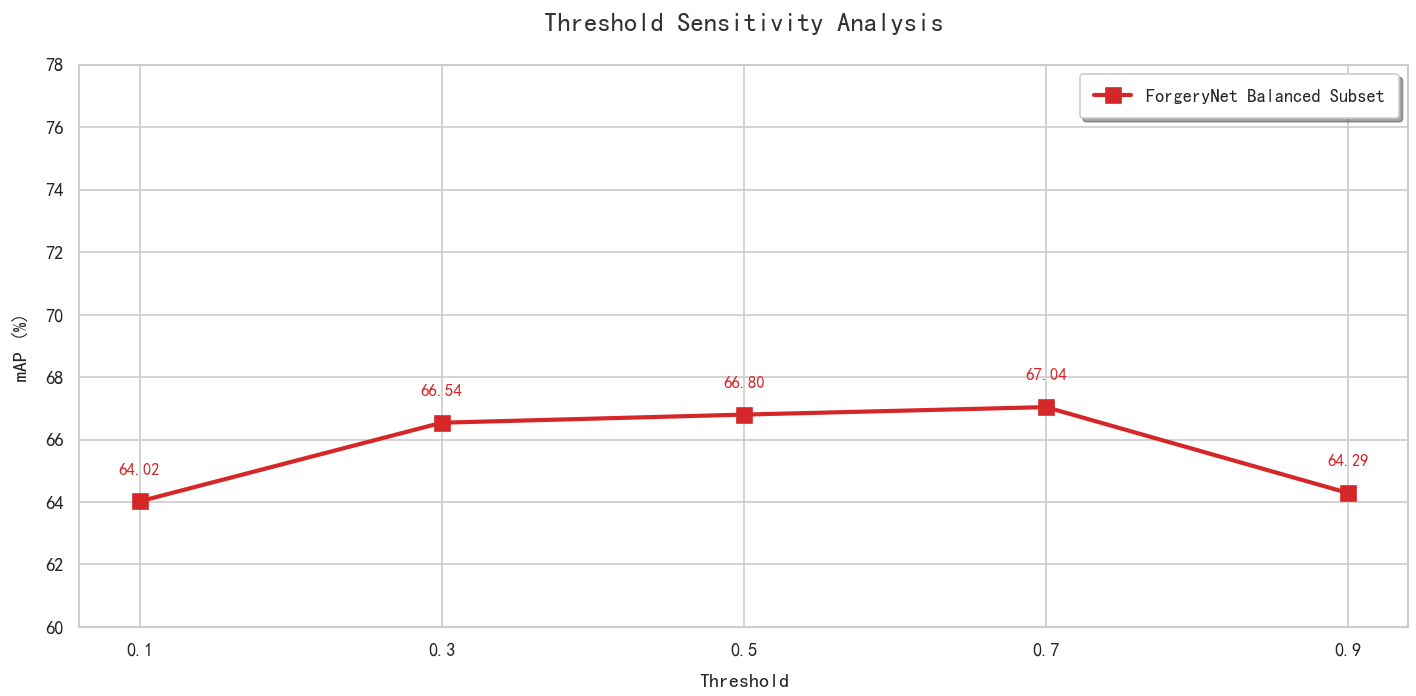}
    \caption{{Sensitivity Analysis of $\tau$.} The curves illustrate the trade-off between noise suppression and semantic preservation. The mAP is calculated as the average of AP@0.5, AP@0.75, and AP@0.95.}
    \label{fig:tau_curve}
\end{figure}

To determine the optimal value for $\tau$, we evaluated the model's performance on both \emph{ForgeryNet Balanced} subset by varying $\tau$ in the range of $\{0.1, 0.3, 0.5, 0.7, 0.9\}$. The quantitative results are visualized in Fig.~\ref{fig:tau_curve}.

\textbf{Analysis of the trade-off:}
As $\tau$ increases from 0.1 to 0.7, we observe a consistent performance gain, peaking at \textbf{67.04\%}. This trend indicates that the initial feature similarity matrix contains considerable noise, and a moderate threshold is essential to prune spurious edges. However, when $\tau$ is further increased to 0.9, performance degrades, suggesting that an overly high threshold removes critical semantic cues necessary for detecting subtle forgeries.

\textbf{Conclusion:}
On the \emph{ForgeryNet Balanced} subset, we observe that the performance steadily improves as $\tau$ increases, peaking at $\tau=0.7$ (67.04\%). This indicates that $\tau=0.7$ achieves the optimal trade-off between filtering out noise and preserving meaningful semantic dependencies. Consequently, we empirically set $\tau=0.7$ as the default configuration for our final model.


\end{document}